\documentclass[12pt,twoside]{article}
\usepackage{wrapfig}
\usepackage{graphicx}
  \newdimen\paravsp  \paravsp=1.3ex
\newenvironment{keywords}{\centerline{\bf\small
Keywords}\begin{quote}\small}{\par\end{quote}\vskip 1ex}
\def\idx#1{\index{#1}#1} %\idx{name} for also in text
\def\indxs#1#2{\index{#1!#2}\index{#2!#1}} %\idx{name} for also in text
\def\paradot#1{\vspace{\paravsp plus 0.5\paravsp minus 0.5\paravsp}\noindent{\bf\boldmath{#1.}}}
\def\paranodot#1{\vspace{\paravsp plus 0.5\paravsp minus 0.5\paravsp}\noindent{\bf\boldmath{#1}}}
\def\nq{\hspace{-1em}}
\def\emb{\itshape\bfseries}
\def\emi{\itshape}

\begin{document}
%%%%%%%%%%%%%%%%%%%%%%%%%%%%%%%%%%%%%%%%%%%%%%%%%%%%%%%%%%%%%%%
%%                    T i t l e - P a g e                    %%
%%%%%%%%%%%%%%%%%%%%%%%%%%%%%%%%%%%%%%%%%%%%%%%%%%%%%%%%%%%%%%%

\title{
\vskip 2mm\bf\Large\hrule height5pt \vskip 4mm
One Decade of Universal Artificial Intelligence
\vskip 4mm \hrule height2pt}
\author{{\bf Marcus Hutter}\\[3mm]
\begin{minipage}{0.45\textwidth}
\normalsize RSCS$\,$@$\,$ANU and SML$\,$@$\,$NICTA \\
\normalsize Canberra, ACT, 0200, Australia
\end{minipage}
\&~~~~
\begin{minipage}{0.4\textwidth}
\normalsize Department of Computer Science \\
\normalsize ETH Z\"urich, Switzerland
\end{minipage}
}
\date{February 2012}
\maketitle

\begin{abstract}
% One decade of UAI
The first decade of this century has seen the nascency of
the first mathematical theory of general artificial intelligence.
This theory of Universal Artificial Intelligence (UAI)
has made significant contributions to many theoretical,
philosophical, and practical AI questions.
% UAI mathematical foundations
In a series of papers culminating in book (Hutter, 2005), an
exciting sound and complete mathematical model for a super
intelligent agent (AIXI) has been developed and rigorously
analyzed.
% UAI in philosophy (IOR)
While nowadays most AI researchers avoid discussing
intelligence, the award-winning PhD thesis (Legg, 2008)
provided the philosophical embedding and investigated the
UAI-based universal measure of rational intelligence, which is
formal, objective and non-anthropocentric.
% UAI in practice (MC-AIXI-CTW)
Recently, effective approximations of AIXI have been derived and
experimentally investigated in JAIR paper (Veness et al. 2011).
This practical breakthrough has resulted in some impressive
applications, finally muting earlier critique that UAI is only a theory. %
For the first time, without providing any domain knowledge, the
same agent is able to self-adapt to a diverse range of
interactive environments. For instance, AIXI is able to {\em
learn} from scratch to play TicTacToe, Pacman, Kuhn Poker, and
other games by trial and error, without even providing the
rules of the games.

% A new hope. This progress gives ...
These achievements give new hope that the grand goal of
Artificial General Intelligence is not elusive.

% This article
This article provides an informal overview of UAI in context.
It attempts to gently introduce a very theoretical, formal, and
mathematical subject, and discusses philosophical and technical
ingredients, traits of intelligence, some social questions, and
the past and future of UAI.
\newpage\def\contentsname{\centering\normalsize Contents}
{\parskip=-2.7ex\tableofcontents}
\end{abstract}

\begin{keywords}
artificial intelligence;
reinforcement learning;
algorithmic information theory;
sequential decision theory;
universal induction;
rational agents;
foundations.
\end{keywords}

\begin{quote}\it
``The formulation of a problem is often more essential than its solution,
which may be merely a matter of mathematical or experimental skill.
To raise new questions, new possibilities, to regard old problems from a new angle,
requires creative imagination and marks real advance in science.'' \par
\hfill --- {\sl Albert Einstein (1879--1955)}
\end{quote}

%%%%%%%%%%%%%%%%%%%%%%%%%%%%%%%%%%%%%%%%%%%%%%%%%%%%%%%%%%%%%%%
\section{Introduction}\label{sec:Intro}
%%%%%%%%%%%%%%%%%%%%%%%%%%%%%%%%%%%%%%%%%%%%%%%%%%%%%%%%%%%%%%%

\index{artificial intelligence!the dream}%
%-------------------------------%
\paradot{The dream}
%-------------------------------%
% Human Intelligence
\indxs{human}{mind}%
\indxs{human}{identity}%
The {\emi human mind} is one of the great mysteries in the
Universe, and arguably the most interesting phenomenon to
study. After all, it is connected to {\emi consciousness} and
{\emi identity} which define who we are. Indeed, a healthy mind
(and body) is our most precious possession. Intelligence is the
most distinct characteristic of the human mind, and one we are
particularly proud of. It enables us to understand, explore,
and considerably shape our world, including ourselves.
%
% Artificial Intelligence
\index{artificial intelligence}\index{intelligence!artificial}%
The field of {\emi Artificial Intelligence} (AI) is concerned
with the study and construction of artifacts that exhibit
intelligent behavior, commonly by means of computer algorithms.
%
% Artificial General Intelligence
The {\emi\idx{grand goal}} of AI is to develop systems that
exhibit {\emi\idx{general
intelligence}}\index{intelligence!general} on a {\emi
human-level}\indxs{human-level}{intelligence} or beyond.
If achieved, this would have a far greater impact on human society
than all previous inventions together, likely resulting in a
post-human civilization that only faintly resembles current humanity
\cite{Kurzweil:05,Hutter:12singularity}.\indxs{post-human}{civilization}

\index{artificial intelligence!optimists}%
\index{artificial intelligence!pessimists}%
The dream of creating such artificial devices that reach or
outperform our own intelligence is an old one with a persistent
great divide between ``optimists'' and ``pessimists''.
Apart from the overpowering technical challenges, research on
machine intelligence also involves many fundamental
philosophical questions with possibly inconvenient answers:
What is intelligence? Can a machine be intelligent? %
Can a machine have free will? Does a human have free will? %
Is intelligence just an emergent phenomenon of a simple
dynamical system or is it something intrinsically complex? %
What will our ``Mind Children'' be like? %
How does mortality affect decisions and actions? %
to name just a few.

\indxs{artificial intelligence}{history}%
%-------------------------------%
\paradot{What was wrong with last century's AI}
%-------------------------------%
% lack of interest in the foundations
Some claim that AI has not progressed much in the last 50
years. It definitely has progressed much slower than the
fathers of AI expected and/or promised. There are also some
philosophical arguments that the grand goal of creating
super-human AI may even be elusive in principle. Both reasons
have lead to a decreased interest in funding and research on
the foundations of Artificial General Intelligence (AGI).
\indxs{artificial intelligence}{funding}%

% wrong paradigm
\indxs{artificial intelligence}{paradigms}%
The real problem in my opinion is that
early on, AI has focussed on the wrong paradigm, namely
deductive logical; and being unable to get the foundations
right in this framework, AI soon concentrated on practical but
limited algorithms.
% Solomonoff/Cheeseman against Minsky
Some prominent early researchers such as Ray Solomonoff, who
actually participated in the 1956 Dartmouth workshop,
generally regarded as the birth of AI, and later Peter Cheeseman and
others, advocated a probabilistic inductive approach but couldn't compete with
the soon dominating figures such as Marvin Minsky, Nils Nilsson, and others who
advocated a symbolic/logic approach as the foundations of AI.
(of course this paragraph is only a caricature of AI history).

% formal foundations
Indeed it has even become an acceptable attitude that general
intelligence is in principle unamenable to a formal definition.
In my opinion, claiming something to be impossible without
strong evidence sounds close to an unscientific position; and
there {\em are no} convincing arguments against the feasibility of AGI
\cite{Chalmers:96,Legg:08}.

% A(G)I revival
Also, the failure of once-thought-promising AI-paradigms at
best shows that they were not the right approach or maybe they
only lacked sufficient computing power at the time. Indeed,
after early optimism mid-last century followed by an AI
depression, there is renewed, justified, optimism
\cite[Sec.1.3.10]{Russell:10}, as is evident by the new conference
series on Artificial General Intelligence, the Blue Brain
project, the Singularity movement, and
the anthologies \cite{Goertzel:07,Wang:12}
prove.
AI research has come in waves and paradigms (\idx{computation},
\idx{logic}, \idx{expert systems}, \idx{neural nets}, \idx{soft
approaches}, \idx{learning}, \idx{probability}). Finally, with
the free access to unlimited amounts of data on the internet,
{\emi\idx{information}}-centered AI research has blossomed.
\indxs{artificial intelligence}{paradigms}%

\indxs{artificial intelligence}{foundations}%
\indxs{artificial intelligence}{information theory}%
%-------------------------------%
\paradot{New foundations of A(G)I}
%-------------------------------%
Universal Artificial Intelligence (UAI) is such a modern
information-theoretic inductive approach to AGI,
in which logical reasoning plays no direct role.
UAI is a new paradigm to AGI via a path from universal
induction to prediction to decision to action. It has been
investigated in great technical depth \cite{Hutter:04uaibook}
and has already spawned promising formal definitions of
rational intelligence, the optimal rational agent AIXI
and practical approximations thereof, and put AI on solid
mathematical foundations.
It seems that we could, for the first time, have a general
mathematical theory of (rational) intelligence that is sound
and complete in the sense of well-defining the general AI
problem as detailed below.
The theory allows a rigorous mathematical investigation of many
interesting philosophical questions surrounding (artificial)
intelligence. Since the theory is complete, definite answers
can be obtained for a large variety of intelligence-related
questions, as foreshadowed by the award winning PhD thesis of
\cite{Legg:08}.

%-------------------------------%
\paradot{Contents}
%-------------------------------%
Section~\ref{sec:Problem} provides the context and background
for UAI. It will summarize various last century's
paradigms for and approaches to understanding and building
artificial intelligences, highlighting their problems and how
UAI is similar or different to them.
Section~\ref{sec:UAI} then informally describes the ingredients
of UAI. It mentions the UAI-based intelligence measure only in
passing to go directly to the core AIXI definition. In which
sense AIXI is the most intelligent agent and a theoretical
solution of the AI problem is explained.
Section~\ref{sec:Traits} explains how the complex phenomenon of
intelligence with all its facets can emerge from the simple
AIXI equation.
Section~\ref{sec:Social} considers an embodied version of
AIXI embedded into our society. I go through some important
social questions and hint at how AIXI might behave, but this
is essentially unexplored terrain.
The technical state-of-the-art/development of UAI is summarized in Section~\ref{sec:State}: %
theoretical results for AIXI and universal Solomonoff induction; %
practical approximations, implementations, and applications of AIXI; %
UAI-based intelligence measures, tests, and definitions; %
and the human knowledge compression contest.
Section \ref{sec:Disc} concludes with a summary and outlook how
UAI helps in formalizing and answering deep philosophical
questions around AGI and last but not least how to build super
intelligent agents.

%%%%%%%%%%%%%%%%%%%%%%%%%%%%%%%%%%%%%%%%%%%%%%%%%%%%%%%%%%%%%%%
\section{The AGI Problem}\label{sec:Problem}
%%%%%%%%%%%%%%%%%%%%%%%%%%%%%%%%%%%%%%%%%%%%%%%%%%%%%%%%%%%%%%%

\index{the AI problem}%
\indxs{artificial intelligence}{foundations}%
The term AI means different things to different people. I will
first discuss why this is so, and will argue that this due to a
lack of solid and generally agreed-upon foundations of AI. The
field of AI soon abandoned its efforts of rectifying this state
of affairs, and pessimists even created a defense mechanism
denying the possibility or usefulness of a (simple) formal
theory of general intelligence. While human intelligence might
indeed be messy and unintelligible, I will argue that a simple
formal definition of machine intelligence {\em is} possible and
useful.
I will discuss how this definition fits into the various %
important dimensions of research on (artificial) intelligence including %
human$\leftrightarrow$rational, %
thinking$\leftrightarrow$acting, %
top-down$\leftrightarrow$bottom-up, %
the agent framework, %
traits of intelligence, %
deduction$\leftrightarrow$induction, and %
learning$\leftrightarrow$planning. %

\indxs{artificial intelligence}{the problem}%
%------------------------------%
\paradot{The problem}
%------------------------------%
I define {\em\idx{the AI problem}} to mean the problem of
building systems that possess general, rather than specific,
intelligence in the sense of being able to solve a wide range
of problems generally regarded to require human-level
intelligence.

\indxs{artificial intelligence}{optimists}%
\indxs{artificial intelligence}{pessimists}%
\indxs{artificial intelligence}{philosophy}%
% Optimists versus pessimists
Optimists believe that the AI problem can be solved within a
couple of decades \cite{Kurzweil:05}. Pessimists deny its
principle feasibility on religious, philosophical,
mathematical, or technical grounds
(see \cite[Chp.26]{Russell:10} for a list of arguments). Optimists
have refuted/rebutted all those arguments (see \cite[Chp.9]{Chalmers:96}
and \cite{Legg:08}), but haven't produced super-human AI
either, so the issue remains unsettled.

% Vague and not-agreed-upon foundations
One problem in AI, and I will argue key problem, is that
there is no general agreement on what intelligence is. This
has lead to endless circular and often fruitless arguments, and
has held up progress. Generally, the lack of a generally-accepted
solid foundation makes high card houses fold easily. Compare
this with Russell's paradox which shattered the foundations of
mathematics, and which was finally resolved by the
completely formal and generally agreed-upon ZF(C) theory of
sets.

% AI avoids discussing or formalizing intelligence
On the other hand, it is an anomaly that nowadays most AI
researchers avoid discussing or formalizing intelligence, which
is caused by several factors: It is a difficult old subject, it
is politically charged, it is not necessary for narrow AI which
focusses on specific applications, AI research is done primarily
by computer scientists who mainly care about algorithms rather
than philosophical foundations, and the popular belief that
general intelligence is principally unamenable to a
mathematical definition. These reasons explain but only
partially justify the limited effort in trying to formalize
general intelligence. There is no convincing argument that this
is impossible.

% Why a definition of intelligence is important
Assume we had a formal, objective, non-anthropocentric, and
direct definition, measure, and/or test of intelligence, or at
least a very general intelligence-resembling formalism that
could serve as an adequate substitute. This would bring the
higher goals of the field into tight focus and allow us to
objectively and rigorously compare different approaches and
judge the overall progress.
Formalizing and rigorously defining a previously vague concept
usually constitutes a quantum leap forward in the field: Cf.\ the
history of sets, numbers, logic, fluxions/infinitesimals, energy,
infinity, temperature, space, time, observer, etc.

% Is a formal definition of intelligence possible ?
Is a simple {\emi formal definition of intelligence} possible?
Isn't intelligence a too complex and anthropocentric phenomenon
to allow formalization? Likely not: There are very simple
models of chaotic phenomena such as turbulence. Think about the
simple iterative map $z\to z^2+c$ that produces the amazingly
rich, fractal landscape, sophisticated versions of it used to
produce images of virtual ecosystems as in the movie Avatar.
Or the complexity of (bio)chemistry emerges out of the
elegant mathematical theory Quantum Electro Dynamics.

% human intelligence is messy
Modeling human intelligence is probably going to be messy, but
ideal rational behavior seems to capture the essence of
intelligence, and, as I claim, can indeed be completely
formalized.
Even if there is no unique definition capturing all aspects we
want to include in a definition of intelligence, or if some
aspects are forever beyond formalization (maybe consciousness
and qualia), pushing the frontier and studying the best
available formal proxy is of utmost importance for
understanding artificial and natural minds.

\indxs{artificial intelligence}{context}%
%------------------------------%
\paradot{Context}
%------------------------------%
% fields relevant for A(G)I
There are many fields that try to understand the phenomenon of
intelligence and whose insights help in creating intelligent
systems: %
{\emi cognitive \idx{psychology}} \cite{Solso:07} and {\emi\idx{behaviorism}} \cite{Skinner:74}, %
{\emi philosophy of mind}\index{philosophy!of the mind}\index{mind!philosophy} \cite{Chalmers:02,Searle:05}, %
{\emi\idx{neuroscience}} \cite{Hawkins:04}, %
{\emi\idx{linguistics}} \cite{Hausser:01,Chomsky:06}, %
{\emi\idx{anthropology}} \cite{Park:07}, %
{\emi\idx{machine learning}} \cite{Sutton:98,Bishop:06}, %
{\emi\idx{logic}} \cite{Turner:84,Lloyd:87}, %
{\emi\idx{computer science}} \cite{Russell:10}, %
{\emi biological \idx{evolution}} \cite{Tettamanzi:01,Kardong:07}, %
{\emi\idx{economics}} \cite{McKenzie:09}, % (rational decision making)
and {\emi others}.

% human<->rational and thinking<->acting
\begin{wrapfigure}{r}{45ex}\vspace{-3ex}\tabcolsep2pt
\hfill\begin{tabular}{|c|c|c|} \hline
  \em\bf What is AI? & \bf Thinking  & \bf Acting \\ \hline
  \bf humanly & Cognitive & Turing test,      \\[-0.5ex]
              & Science   & Behaviorism       \\ \hline
\bf rationally & Laws of  & {\bf Doing the}   \\[-0.5ex]
              & Thought   & {\bf Right Thing} \\ \hline
\end{tabular}\vspace{-3ex}
\end{wrapfigure}
\indxs{rational}{action}%
\indxs{rational}{thinking}%
\indxs{human}{action}%
\indxs{human}{thinking}%
\index{cognitive science}%
\indxs{Turing}{test}%
\index{laws of thought}%
Cognitive science studies how humans think, %
Behaviorism and the Turing test how humans act, %
the laws of thought define rational thinking, %
while AI research increasingly focusses
on systems that act rationally.

% Bottom-up versus top-down
In computer science, most AI research is {\emi\idx{bottom-up}};
extending and improving existing or developing new {\em
algorithms} and increasing their range of applicability; an
interplay between experimentation on toy problems and theory,
with occasional real-world applications. A {\emi\idx{top-down}}
approach would start from a general principle and derive
effective approximations (like heuristic approximations to
minimax tree search).
Maybe when the top-down and bottom-up approaches meet in the
middle, we will have arrived at practical truly intelligent
machines.

% Agent framework
\indxs{agent}{framework}%
\indxs{artificial}{agent}%
The science of artificial intelligence may be defined as the
construction of intelligent systems ({\emi artificial agents})
and their analysis. A natural definition of a {\emi\idx{system}} is
anything that has an input and an output stream, or
equivalently an agent that acts and observes. This agent
perspective of AI \cite{Russell:10} brings some order and
unification into the large variety of problems the fields wants
to address, but it is only a framework rather than providing a
complete theory of intelligence.
\index{machine learning}%
In the absence of a perfect (stochastic) model of the
environment the agent interacts with, {\emi\idx{machine learning}}
techniques are needed and employed to learn from experience.
There is no general theory for learning agents (apart from UAI).
This has resulted in an ever increasing number of {\emi limited
models and algorithms} in the past.

% Facets/traits of intelligence
What distinguishes an {\em intelligent} system from a
non-intelligent one? {\emi Intelligence} can have many faces
like {\em reasoning}, {\em creativity}, {\em association}, {\em
generalization}, {\em pattern recognition}, {\em problem
solving}, {\em memorization}, {\em planning}, {\em achieving
goals}, {\em learning}, {\em optimization}, {\em
self-preservation}, {\em vision}, {\em language processing},
{\em classification}, {\em induction}, {\em deduction}, and
{\em knowledge acquisition and processing}. A formal definition
incorporating every aspect of intelligence, however, {\em seems}
difficult.

% Informal definitions of intelligence
\indxs{intelligence}{definition}%
There is no lack of attempts to characterize or define
intelligence trying to capture all traits {\em informally}
\cite{Hutter:07idefs}. One of the more successful
characterizations is: {\emi Intelligence measures an agents
ability to perform well in a large range of environments}
\cite{Hutter:07iorx}. Most traits of intelligence are implicit
in and emergent from this definition as these capacities enable
an agent to succeed \cite{Legg:08}. Convincing
formal definitions other than the ones spawned by UAI are
essentially lacking.

% Induction versus deduction
Another important dichotomy is whether an approach focusses
(more) on \idx{deduction} or \idx{induction}. Traditional AI
concentrates mostly on the logical deductive reasoning aspect,
while machine learning focusses on the inductive inference
aspect. Learning and hence induction are indispensable
traits of any AGI. Regrettably, induction is peripheral to
traditional AI, and the machine learning community in large is
not interested in A(G)I. It is the field of \idx{reinforcement
learning} at the intersection of AI and machine learning that
has AGI ambitions {\em and} takes learning seriously.

%-------------------------------%
\paradot{UAI in perspective}
%-------------------------------%
The theory of Universal Artificial Intelligence developed in
the last decade is a modern information-theoretic, inductive,
reinforcement learning approach to AGI that has been
investigated in great technical depth \cite{Hutter:04uaibook}.

% field of AI, top-down, agent, not framework, traits, formal, general
Like traditional AI, UAI is concerned with agents {\em doing
the right thing}, but is otherwise quite different: It is a {\em
\idx{top-down}} approach in the sense that it starts with a single
completely {\em formal} {\em general} definition of
intelligence from which an essentially {\em unique agent} that
seems to possess all {\em traits} of rational intelligence is
derived. It is not just another framework with some gaps to be
filled in later, since the agent is {\em completely} defined.

% induction rather than logical deduction
\indxs{universal}{learning}%
It also takes induction very seriously: Universal learning is one
of the agent's two key elements (the other is stochastic
planning). Indeed, logic and deduction play no fundamental role
in UAI (but are emergent).
This also naturally dissolves Lucas' and Penrose'
\cite{Penrose:94} argument against AGI that Goedel's
incompleteness result shows that the human mind is not a
computer. The fallacy is to assume that the mind (human and
machine alike) are infallible deductive machines.

% Status of UAI<->String Theory
The status of UAI might be compared to \idx{Super String
theory} in physics. Both are currently the most promising
candidates for a \idx{grand unification} (of AI and physics,
respectively), although there are also marked differences. Like
the unification hierarchy of physical theories allows relating
and regarding the myriad of limited models as effective
approximations, UAI allows us to regard existing approaches to
AI as effective approximations. %
Understanding AI in this way gives researchers a much more
coherent view of the field.

% transition to next section
Indeed, UAI seems to be the first sound and complete
mathematical theory of (rational) intelligence. The next
section presents a very brief introduction to UAI from
\cite{Hutter:09aixiopen}, together with an informal explanation
of what the previous sentence actually means. See
\cite{Hutter:04uaibook} for formal definitions and results.

%%%%%%%%%%%%%%%%%%%%%%%%%%%%%%%%%%%%%%%%%%%%%%%%%%%%%%%%%%%%%%%
\section{Universal Artificial Intelligence}\label{sec:UAI}
%%%%%%%%%%%%%%%%%%%%%%%%%%%%%%%%%%%%%%%%%%%%%%%%%%%%%%%%%%%%%%%

This section describes the theory of Universal Artificial
Intelligence (UAI), a modern information-theoretic approach to
AI, which differs essentially from mainstream A(G)I research
described in the previous sections.
The connection of UAI to other research fields and the
philosophical and technical ingredients of UAI (Ockham,
Epicurus, Turing, Bayes, Solomonoff, Kolmogorov, Bellman) are
briefly discussed.
The UAI-based universal intelligence measure and order relation
in turn define the (w.r.t.\ this measure) most intelligent
agent AIXI, which seems to be the first sound and complete
theory of a universal optimal rational agent embedded in an
arbitrary computable but unknown environment with reinforcement
feedback. The final paragraph clarifies what this actually means.

\index{intelligence!formal definition}%
\index{definition!intelligence}%
%------------------------------%
\paradot{Defining Intelligence}
%------------------------------%
Philosophers, AI researchers, psychologists, and others have
suggested many informal=verbal definitions of intelligence
\cite{Hutter:07idefs}, but there is not too much work on formal
definitions that are broad, objective, and non-anthropocentric.
See \cite{Hutter:07iorx} for a comprehensive collection,
discussion and comparison of intelligence definitions, tests,
and measures with all relevant references.
It is beyond the scope of this article to discuss them.

\index{intelligence!measure}%
\index{intelligence!order relation}%
% universal intelligence measure (UIM) / intelligence order relation (IOR)
Intelligence is graded, since agents can be more or less
intelligent. Therefore it is more natural to consider measures
of intelligence, rather than binary definitions which would
classify agents as intelligent or not based on an (arbitrary)
threshold. This is exactly what UAI provides: A formal, broad,
objective, universal measure of intelligence
\cite{Hutter:07iorx}, which formalizes the verbal
characterization stated in the previous section. Agents can be
more or less intelligent w.r.t.\ this measure and hence can be
sorted w.r.t.\ their intelligence
\cite[Sec.5.1.4]{Hutter:04uaibook}. One can show that there is
an agent, coined AIXI, that maximizes this measure, which could
therefore be called the most intelligent agent.

I will not present the UAI-based intelligence measure
\cite{Hutter:07iorx} and order relation \cite{Hutter:04uaibook}
here, but, after listing the conceptual ingredients to UAI and
AIXI, directly proceed to defining and discussing AIXI.

\indxs{universal}{artificial intelligence}%
%------------------------------%
\paradot{UAI and AIXI ingredients \cite{Hutter:09aixiopen}}
%------------------------------%
The theory of UAI has interconnections with (draws from and
contributes to)
many research fields, encompassing %
computer science (artificial intelligence, machine learning, computation), %
engineering (information theory, adaptive control), %
economics (rational agents, game theory), %
mathematics (statistics, probability), %
psychology (behaviorism, motivation, incentives), and %
philosophy (inductive inference, theory of knowledge). %
The concrete ingredients in AIXI are as follows: Intelligent
{\emi\idx{actions}} are based on informed
{\emi\idx{decisions}}. Attaining good decisions requires
{\emi\idx{predictions}} which are typically based on models of
the environments. Models are constructed or learned from past
observations via {\emi\idx{induction}}.
Fortunately, based on the {\emi deep philosophical insights}
and {\emi powerful mathematical developments}, all these
problems have been overcome, at least in theory:
So what do we need (from a mathematical point of view) to
construct a universal optimal learning agent interacting with
an arbitrary unknown environment? The theory, coined {\emi
UAI}, developed in the last decade and explained in
\cite{Hutter:04uaibook} says: {\emi All you need is}
{\em\idx{Ockham}}, {\em\idx{Epicurus}}, {\em\idx{Turing}},
{\em\idx{Bayes}}, {\em\idx{Solomonoff}} \cite{Solomonoff:64},
{\em\idx{Kolmogorov}} \cite{Kolmogorov:65}, and {\em\idx{Bellman}}
\cite{Bellman:57}: %
Sequential \idx{decision theory} \cite{Bertsekas:06} ({\emi
Bellman}'s equation) formally solves the problem of rational
agents in uncertain worlds if the true environmental
probability distribution is known. If the environment is
unknown, {\emi Bayes}ians \cite{Berger:93} replace the true
distribution by a weighted mixture of distributions from some
(hypothesis) class. Using the large class of all (semi)measures
that are (semi)computable on a {\emi Turing} machine bears in
mind {\emi Epicurus}, who teaches not to discard any
(consistent) hypothesis. In order not to ignore {\emi Ockham},
who would select the simplest hypothesis, {\emi Solomonoff}
defined a universal prior that assigns high/low prior weight to
simple/complex environments \cite{Hutter:11uiphil}, where {\emi
Kolmogorov} quantifies complexity \cite{Li:08}.
Their unification constitutes the theory of UAI and resulted in
the universal intelligence measure and order relation and the following
model/agent AIXI.

\index{AIXI!definition}%
%------------------------------%
\paradot{The AIXI Model in one line \cite{Hutter:09aixiopen}}
%------------------------------%
%
\def\nc{}% normal color (black)
\def\ac{}% agent color
\def\oc{}% observation color
\def\rc{}% reward color
\def\xc{}% perception color
\def\ec{}% environment color
\def\uc{}% universal color
\def\ic{}% AIXI color
It is possible to write down the {\ic AIXI} model explicitly in
one line, although {\em one should not expect to be able to
grasp the full meaning and power from this compact and somewhat
simplified representation}.

\begin{wrapfigure}{r}{41ex}
  \small\boldmath
  \unitlength=0.4ex
\linethickness{0.4pt}
\hspace*{2ex}\begin{picture}(106,34)(0,9) %(106,47)
\thinlines
\put( 1,41){\framebox(16,6)[cc]{${\rc r_1}|{\oc o_1}$}}
\put(17,41){\framebox(16,6)[cc]{${\rc r_2}|{\oc o_2}$}}
\put(33,41){\framebox(16,6)[cc]{${\rc r_3}|{\oc o_3}$}}
\put(49,41){\framebox(16,6)[cc]{${\rc r_4}|{\oc o_4}$}}
\put(65,41){\framebox(16,6)[cc]{${\rc r_5}|{\oc o_5}$}}
\put(81,41){\framebox(16,6)[cc]{${\rc r_6}|{\oc o_6}$}}
\put(97,47){\line(1,0){9}}\put(97,41){\line(1,0){9}}\put(102,44){\makebox(0,0)[cc]{\ec...}}
\put( 1,1){\framebox(16,6)[cc]{$\ac a_1$}}
\put(17,1){\framebox(16,6)[cc]{$\ac a_2$}}
\put(33,1){\framebox(16,6)[cc]{$\ac a_3$}}
\put(49,1){\framebox(16,6)[cc]{$\ac a_4$}}
\put(65,1){\framebox(16,6)[cc]{$\ac a_5$}}
\put(81,1){\framebox(16,6)[cc]{$\ac a_6$}}
\put(97,7){\line(1,0){9}}\put(97,1){\line(1,0){9}}\put(102,4){\makebox(0,0)[cc]{\ac...}}
\ac
\put(1,21){\framebox(16,6)[cc]{work}}
\thicklines
\put(17,17){\framebox(20,14)[cc]{\sf Agent}}
\thinlines
\put(37,27){\line(1,0){14}}
\put(37,21){\line(1,0){14}}
\put(39,24){\makebox(0,0)[lc]{tape ...}}
\ec
\put(56,21){\framebox(16,6)[cc]{work}}
\thicklines
\put(72,17){\framebox(20,14)[cc]{}}
\put(82,24){\makebox(0,0)[cb]{\footnotesize\sf Environ-}}
\put(82,24){\makebox(0,0)[ct]{\footnotesize\sf ment}}
\thinlines
\put(92,27){\line(1,0){14}}
\put(92,21){\line(1,0){14}}
\put(94,24){\makebox(0,0)[lc]{tape ...}}
\normalcolor
\thicklines
\put(46,41){\vector(-2,-1){20}}
\put(81,31){\vector(-2,1){20}}
\put(54,7){\vector(3,1){30}}
\put(24,17){\vector(3,-1){30}}
\end{picture}
\end{wrapfigure}
.$\!${\ic AIXI} is an {\ac\idx{agent}} that interacts with an
{\ec\idx{environment}} in cycles $k=1,2,...,m$. In cycle $k$,
{\ic AIXI} takes {\ac\idx{action} $a_k$} (e.g.\ a limb movement)
based on past \idx{perception}s $\oc o_1\rc r_1..\oc o_{k-1}\rc
r_{k-1}$ as defined below.
There\-after, the {\ec environment} provides a (regular)
{\oc\idx{observation} $o_k$} (e.g.\ a camera image) to {\ic AIXI}
and a real-valued {\rc\idx{reward} $r_k$}. The {\rc reward} can be
very scarce, e.g.\ just +1 (-1) for winning (losing) a chess
game, and 0 at all other times. Then the next cycle $k+1$
starts.
This agent-environment interaction protocol can be depicted as
on the right. Given the interaction protocol above, the
simplest version of AIXI is defined by
$$
  \mbox{\ic AIXI}\qquad \ac a_k \nc\;:=\; \ac\arg\max_{a_k}\xc\sum_{\oc o_k\rc r_k} %\max_{a_{k+1}}\sum_{x_{k+1}}
  \nc ... \ac\max_{a_m}\xc\sum_{\oc o_m\rc r_m}
  \rc[r_k+...+r_m]
  \uc\nq\nq\nq\!\!\!\sum_{{\ec q}\,:\,U({\ec q},{\ac a_1..a_m})={\oc o_1\rc r_1..\oc o_m\rc r_m}}\nq\nq\nq\!\!\! 2^{-\ell({\ec q})}
$$
The expression shows that {\ic AIXI} tries to {\ac max}imize
its {\rc total future reward $r_k+...+r_m$}. If the environment
is modeled by a deterministic {\ec program $q$}, then the
future {\xc perceptions} $\rc...{\oc o_k\rc r_k..\oc o_m\rc
r_m} = \uc U({\ec q},{\ac a_1..a_m})$ can be computed, where
$\uc U$ is a {\uc universal (monotone Turing) machine}
executing $\ec q$ given $\ac a_1..a_m$. Since $\ec q$ is
unknown, {\ic AIXI} has to maximize its {\xc expected} reward,
i.e.\ average $\rc r_k+...+r_m$ over all possible future {\ec
perceptions} created by all possible environments $\ec q$ that
are consistent with past perceptions.
The simpler an environment, the higher is its a-priori
contribution $\uc 2^{-\ell({\ec q})}$, where simplicity is
measured by the {\uc length $\ell$} of program $\ec q$.
{\ic AIXI} effectively learns by eliminating Turing machines $\ec q$
once they become inconsistent with the progressing \idx{history}.
\indxs{noisy}{environment}%
\indxs{deterministic}{environment}%
\indxs{mixture}{environment}%
\indxs{chronological}{environment}%
Since {\ec noisy environments} are just mixtures of
deterministic environments, they are automatically included
\cite[Sec.7.2]{Hutter:11uiphil},\cite{Hutter:11unipreq}.
The sums in the formula constitute the averaging process. {\xc
Averaging} and {\ac maximization} have to be performed in
chronological order, hence the interleaving of {\ac max} and
$\xc\Sigma$ (similarly to minimax for games).

One can fix any finite action and perception space, any
reasonable $\uc U$, and any large finite lifetime $m$. This
completely and uniquely defines {\ic AIXI}'s actions $\ac a_k$,
which are limit-computable via the expression above (all quantities
are known).

%------------------------------%
\paradot{Discussion}
%------------------------------%
% Explanation of sound, complete, universal, optimal, rational, theory.
The AIXI model seems to be the first sound and complete {\em
theory} of a universal optimal rational agent embedded in an
arbitrary computable but unknown environment with reinforcement
feedback.
\indxs{universal}{AIXI}%
AIXI is {\emi universal} in the sense that it is designed to be able
to interact with any (deterministic or stochastic) computable
environment; the universal Turing machines on which it
is based is crucially responsible for this.
\indxs{complete}{AIXI}%
AIXI is {\emi complete} in the sense that it is not an
incomplete framework or partial specification (like Bayesian
statistics which leaves open the choice of the prior or the
rational agent framework or the subjective expected
utility principle) but is completely and essentially
uniquely defined.
\indxs{sound}{AIXI}%
AIXI is {\emi sound} in the sense of being (by construction)
free of any internal contradictions (unlike e.g.\ in
knowledge-based deductive reasoning systems where avoiding
inconsistencies can be very challenging).
\indxs{optimal}{AIXI}%
AIXI is {\emi optimal} in the senses that: no other agent can
perform uniformly better or equal in all environments, it is a
unification of two optimal theories themselves, a variant is
self-optimizing; and it is likely also optimal in other/stronger
senses.
\indxs{rational}{AIXI}%
AIXI is {\emi rational} in the sense of trying to maximize its
future long-term reward.
\indxs{theory}{AIXI}%
% explaining AIXI solves AI
For the reasons above I have argued that AIXI is a
mathematical ``solution'' of the AI problem: AIXI would be able
to learn any learnable task and likely better so than any other
unbiased agent, but AIXI is more a {\emi theory} or formal
definition rather than an algorithm, since it is only
limit-computable.
%
% How can 1 line capture intelligence
How can an equation that fits into a single line capture the
diversity, complexity, and essence of (rational) intelligence?
We know that complex appearing phenomena such as chaos and
fractals can have simple descriptions such as iterative maps and
the complexity of chemistry emerges from simple physical laws.
There is no a-priori reason why ideal rational intelligent
behavior should not also have a simple description, with
most traits of intelligence being emergent.
% Axiomatic characterization
Indeed, even an axiomatic characterization seems possible
\cite{Hutter:11aixiaxiom,Hutter:11aixiaxiom2}.

%%%%%%%%%%%%%%%%%%%%%%%%%%%%%%%%%%%%%%%%%%%%%%%%%%%%%%%%%%%%%%%
\section{Facets of Intelligence}\label{sec:Traits}
%%%%%%%%%%%%%%%%%%%%%%%%%%%%%%%%%%%%%%%%%%%%%%%%%%%%%%%%%%%%%%%

\indxs{intelligence}{facets}%
\indxs{emergent}{intelligence}%
Intelligence can have many faces. I will argue in this section
that the AIXI model possesses all or at least most properties
an intelligent rational agent should possess. Some facets have
already been formalized, some are essentially built-in, but the
majority have to be emergent. Some of the claims have been
proven in \cite{Hutter:04uaibook} but the majority has yet to
be addressed.

%------------------------------%
\paranodot{Generalization}
%------------------------------%
\index{generalization}\index{induction}%
is essentially inductive inference \cite{Hutter:11uiphil}.
{\emb Induction} is the process of inferring general laws or
models from observations or data by finding regularities in
past/other data. This trait is a fundamental cornerstone of
intelligence.

%------------------------------%
\paranodot{Prediction}
%------------------------------%
\index{prediction}%
is concerned with forecasting future
observations (often based on models of the world learned)
from past observations. Solomonoff's theory of prediction
\cite{Solomonoff:64,Solomonoff:78} is a universally optimal
solution of the prediction problem
\cite{Hutter:07uspx,Hutter:11uiphil}.
Since it is a key ingredient in the AIXI model, it is
natural to expect that AIXI is an optimal predictor if
rewarded for correct predictions. Curiously only weak and
limited rigorous results could be proven so far
\cite[Sec.6.2]{Hutter:04uaibook}.

%------------------------------%
\paranodot{Pattern recognition},
%------------------------------%
\index{pattern recognition}%
abstractly speaking, is concerned with classifying data
(patterns). This requires a similarity measure between
patterns. Supervised {\emb\idx{classification}} can essentially
be reduced to a sequence prediction problem, hence formally
pattern recognition reduces to the previous item, although
interesting questions specific to classification emerge
\cite[Chp.3]{Hutter:04uaibook}.

%------------------------------%
\paradot{Association}
%------------------------------%
\index{association}\index{clustering}%
Two stimuli or observations are associated if there exists some
(cor)relation between them. A set of observations can often be
{\emb clustered} into different categories of
similar=associated items. For AGI, a {\em universal} similarity
measure is required. Kolmogorov complexity via the universal
similarity metric \cite{Cilibrasi:05} can provide such a measure,
but many fundamental questions have yet to be explored: How
does association function in AIXI? How can Kolmogorov
complexity well-define the (inherently? so far?) ill-defined
clustering problem?

%------------------------------%
\paranodot{Reasoning}
%------------------------------%
\index{reasoning}%
is arguably the most prominent trait of human intelligence.
Interestingly deductive reasoning and logic are {\emb not} part
of the AIXI architecture. The fundamental assumption is that
there is no sure knowledge of the world, all inference is
tentative and inductive, and that logic and {\emb\idx{deduction}}
constitute an idealized limit applicable in situations where
uncertainties are extremely small, i.e.\ probabilities are
extremely close to 1 or 0. What would be very interesting to
show is that {\emb\idx{logic}} is an emergent phenomenon, i.e.\ that
AIXI learns to reason logically if/since this helps collect
reward.

%------------------------------%
\paranodot{Problem solving}
%------------------------------%
\index{problem solving}%
might be defined as goal-oriented reasoning, and hence reduces to
the previous item, since AIXI is designed to {\emb achieve
\idx{goals}} (which is reward maximization in the special case of a
terminal reward when the goal is achieved). Problems can be of
very different nature, and some of the other traits of
intelligence can be regarded as instances of problem solving,
e.g.\ planning.

%------------------------------%
\paranodot{Planning}
%------------------------------%
\index{planning}%
ability is directly incorporated in AIXI via the alternating
maximization and summation in the definition. Algorithmically
AIXI plans through its entire life via a deep expectimax tree
search up to its death, based on its belief about the world. In
known constrained domains this search corresponds to classical
exact planning strategies as e.g.\ exemplified in
\cite[Chp.6]{Hutter:04uaibook}.

%------------------------------%
\paranodot{Creativity}
%------------------------------%
\index{creativity}%
is the ability to generate innovative ideas and
to manifest these into reality. Creative people are often more
successful than unimaginative ones. Since AIXI is the ultimate
success-driven agent, AIXI should be highly creative, but this has
yet to be formalized and proven, or at least exemplified.

%------------------------------%
\paradot{Knowledge}
%------------------------------%
\index{knowledge}%
AIXI stores the entire interaction history and has perfect
{\emb \idx{memory}}. Additionally, models of the experienced
world are constructed (learned) from this
{\emb\idx{information}} in form of short(est) programs. These
models guide AIXI's behavior, so constitute knowledge for AIXI.
Any {\emb\idx{ontology}} is implicit in these programs.
How short-term, long-term, relational, hierarchical,
etc.\ memory emerges out of this compression-based approach has
not yet been explored.

%------------------------------%
\paranodot{Actions}
%------------------------------%
\index{actions}\index{decisions}%
influence the environment which reacts back to the agent. {\emb
Decisions} can have long-term consequences, which the
expectimax planner of AIXI should properly take into account.
Particular issues of concern are the interplay of learning and
planning (the infamous exploration$\leftrightarrow$exploitation
tradeoff \cite{Hutter:11asyoptag}). Additional complications
that arise from embodied agents will be considered in the next
section.

%------------------------------%
\paradot{Learning}
%------------------------------%
\index{learning}%
\index{supervised learning}%
\index{reinforcement learning}%
There are many different forms of learning: supervised,
unsupervised, semi-supervised, reinforcement, transfer,
associative, transductive, prequential, and many others. By
design, AIXI is a reinforcement learner, but one can show that
it will also ``listen'' to an informative teacher, i.e.\ it
{\em learns} to learn supervised \cite[Sec.6.5]{Hutter:04uaibook}.
It is plausible that AIXI can also acquire the other learning
techniques.

%------------------------------%
\paranodot{Self-awareness}
%------------------------------%
\index{self-awareness}%
allows one to (meta)reason about one's own thoughts, which is an
important trait of higher intelligence, in particularly when
interacting with other forms of intelligence. Technically all what
might be needed is that an agent has and exploits not only a model of the
world but also a model of itself including aspects of its own
algorithm, and this recursively. Is AIXI self-aware in this
technical sense?

%------------------------------%
\paranodot{Consciousness}
%------------------------------%
\index{consciousness}%
is possibly the most mysterious trait of the human mind.
Whether anything rigorous can ever be said about the
consciousness of AIXI or AIs in general is not clear and in any
case beyond my expertise. I leave this to philosophers of the
mind \cite{Chalmers:02} like the world-renowned expert on (the
hard problem of) consciousness, David Chalmers
\cite{Chalmers:96}.

%%%%%%%%%%%%%%%%%%%%%%%%%%%%%%%%%%%%%%%%%%%%%%%%%%%%%%%%%%%%%%%
\section{Social Questions}\label{sec:Social}
%%%%%%%%%%%%%%%%%%%%%%%%%%%%%%%%%%%%%%%%%%%%%%%%%%%%%%%%%%%%%%%
\index{social questions}\index{artificial intelligence!social questions}%
\index{artificial intelligence!embodied}\index{embodiment}%
Consider now a sophisticated physical humanoid robot like
Honda's ASIMO but equipped with an AIXI brain. The observations
$o_k$ consist of camera image, microphone signal, and other
\idx{sensory input}. The actions $a_k$ consist of controlling mainly
a loud speaker and motors for limbs, but possibly other
internal functions it has direct control over. The reward $r_k$
should be some combination of its own ``well-being'' (e.g.\
proportional to its battery level and condition of its body
parts) and external reward/punishment from some ``teacher(s)''.

Imagine now what happens if this AIXI-robot is let loose in our
society. Many questions deserving attention arise, and some are
imperative to be rigorously investigated before risking this experiment.

\indxs{nurturing}{environment}%
Children of higher animals require extensive nurturing in a
safe environment because they lack sufficient innate skills for
survival in the real world, but are compensated for their
ability to learn to perform well in a large range of
environments. AIXI is at the extreme of being ``born'' with
\indxs{a-priori}{knowledge}%
essentially no knowledge about our world, but a universal
``brain'' for learning and planning in any environment where
this is possible. As such, it also requires a guiding
\idx{teacher} initially. Otherwise it would simply run out of
battery.

% interface and teaching
AIXI has to learn {\emi\idx{vision}}, {\emi\idx{language}}, and
{\emi\idx{motor skills}} from scratch, similarly to
higher animals and machine learning algorithms, but more
extreme/general. Indeed, Solomonoff \cite{Solomonoff:64} already
showed how his system can learn grammar from positive instances
only, but much remains to be done.
\indxs{training}{sequence}\indxs{reward}{shaping}%
Appropriate {\emi training sequences} and {\emi reward shaping}
in this early ``\idx{childhood}'' phase of AIXI are
important. AIXI can learn from rather crude teachers as
long as the reward is biased in the `right' direction.
The answers to many of the following questions
likely depend on the upbringing of AIXI:

% List of social questions
\begin{itemize}\parskip=0ex\parsep=0ex\itemsep=0ex
\item {\emb Schooling:}\index{schooling} Will a pure reward maximizer
such as AIXI listen to and trust a teacher and learn to learn
supervised (=faster)? Yes \cite[Sec.6.5]{Hutter:04uaibook}.

\item Take {\emb Drugs}\index{drugs} (hacking the reward system):
Likely no, since long-term reward would be small (death), but see
\cite{Ring:11}.

\item {\emb Replication or procreation:}\index{procreation}
Likely yes, if AIXI believes that clones or descendants are useful for its own goals.

\item {\emb Suicide:}\index{suicide} Likely yes (no),
if AIXI is raised to believe to go to heaven (hell)
i.e.\ maximal (minimal) reward forever.

\item {\emb Self-Improvement:}\index{self-improvement}
Likely yes, since this helps to increase reward.

\item {\emb Manipulation:}\index{manipulation}
Manipulate or threaten teacher to give more reward.

\item {\emb Attitude:}\index{attitude} Are pure reward maximizers egoists,
{\emi psychopaths}\index{psychopath}, and/or killers or will they be
{\emi friendly}\indxs{friendly}{artificial intelligence}
({\emi\idx{altruism}} as extended
{\emi\idx{ego(t)ism}})?

\item {\emb Curiosity}\index{curiosity} killed the cat and maybe
AIXI, or is extra reward for curiosity necessary
\cite{Schmidhuber:07curiosity,Orseau:10}?

\item {\emb Immortality}\index{immortality} can cause laziness
\cite[Sec.5.7]{Hutter:04uaibook}!

\item Can {\emb self-preservation}\index{self-preservation} be
learned or need (parts of) it be innate.

\item {\emb Socializing:}\index{socializing} How will AIXI interact
with another AIXI \cite[Sec.5j]{Hutter:09aixiopen},\cite{Hutter:06aixifoe}?
\end{itemize}
A partial discussion of some of these questions can be found in
\cite{Hutter:04uaibook} but many are essentially unexplored.
\indxs{social behavior}{super-intelligence}%
Point is that since AIXI is completely formal, it permits to
formalize these questions and to mathematically analyze them.
That is, UAI has the potential to arrive at definite answers to
various questions regarding the social behavior of
super-intelligences.
Some formalizations and semi-formal answers have recently appeared
in the award-winning papers \cite{Orseau:11agi,Ring:11}.

%%%%%%%%%%%%%%%%%%%%%%%%%%%%%%%%%%%%%%%%%%%%%%%%%%%%%%%%%%%%%%%
\section{State of the Art}\label{sec:State}
%%%%%%%%%%%%%%%%%%%%%%%%%%%%%%%%%%%%%%%%%%%%%%%%%%%%%%%%%%%%%%%

\indxs{UAI}{state of the art}%
This section describes the technical achievements of UAI to
date. Some remarkable and surprising results have already been
obtained. Various theoretical consistency and optimality
results for AIXI have been proven, although stronger results would
be desirable.
On the other hand, the special case of universal induction and
prediction in non-reactive environments is essentially closed.
From the practical side, various computable approximations of
AIXI have been developed, with the latest MC-AIXI-CTW incarnation
exhibiting impressive performance.
Practical approximations of the universal intelligence measure
have also been used to test and consistently order systems of
limited intelligence.
Some other related work such as the compression contest is also
briefly mentioned, and references to some more practical but
less general work such as feature reinforcement learning are given.

\indxs{UAI}{theory}%
\indxs{rational}{intelligence}%
%------------------------------%
\paradot{Theory of UAI}
%------------------------------%
Forceful theoretical arguments that AIXI is the most
intelligent general-purpose agent incorporating all aspects of
rational intelligence have been put forward, supported by
partial proofs.
For this, results of many fields had to be pulled together
or developed in the first place: %
{\emi Kolmogorov complexity}\indxs{Kolmogorov}{complexity} \cite{Li:08},
{\emi\idx{information theory}} \cite{Cover:06},
{\emi sequential \idx{decision theory}} \cite{Bertsekas:06},
{\emi\idx{reinforcement learning}} \cite{Sutton:98},
{\emi\idx{artificial intelligence}} \cite{Russell:10},
{\emi Bayesian statistics}\indxs{Bayesian}{statistics} \cite{Berger:06},
{\emi universal induction}\indxs{universal}{induction} \cite{Hutter:11uiphil}, and
{\emi rational agents}\indxs{rational}{agent} \cite{Shoham:09}. % \cite{Weiss:00},
Various notions of optimality have been considered.
The difficulty is coming up with sufficiently strong
but still satisfiable notions. Some are weaker than desirable,
others are too strong for any agent to achieve.
\indxs{optimal}{AIXI}%
What has been shown thus far is that AIXI learns the correct predictive model \cite{Hutter:04uaibook}, %
is Pareto optimal in the sense that no other agent can
perform uniformly better or equal in all environments, %
and a variant is self-optimizing in the sense that asymptotically
the accumulated reward is as high as possible, i.e.\ the same
as the maximal reward achievable by a completely informed agent \cite{Hutter:02selfopt}. %
AIXI is likely also optimal in other/stronger senses. %
An axiomatic characterization has also been developed
\cite{Hutter:11aixiaxiom,Hutter:11aixiaxiom2}.

%------------------------------%
\paradot{The induction problem}
%------------------------------%
\indxs{induction}{problem}%
The induction problem is a fundamental problem in philosophy
\cite{Earman:93,Hutter:11uiphil} and science
\cite{Jaynes:03,Wallace:05,Gabbay:11}, and a key sub-component
of UAI.
Classical open problems around induction are %
the zero prior problem and the confirmation of
(universal) hypotheses in general and the Black
ravens paradox in particular,
reparametrization invariance,
the old-evidence problem and ad-hoc hypotheses,
and the updating problem \cite{Earman:93}.
In a series of papers (see \cite{Hutter:07uspx} for references)
it has been shown that Solomonoff's theory of universal
induction essentially solves or circumvents all these problems
\cite{Hutter:11uiphil}. It is also predictively optimal and has
minimal regret for arbitrary loss functions.

\indxs{universal}{induction}%
It is fair to say that Solomonoff's theory serves as an
adequate mathematical/theoretical foundation of %
induction \cite{Hutter:11uiphil}, %
machine learning \cite{Hutter:11unilearn}, %
and component of UAI \cite{Hutter:04uaibook}.

%------------------------------%
\paradot{Computable approximations of AIXI}
%------------------------------%
\indxs{approximation}{AIXI}%
An early critique of UAI was that AIXI is incomputable.
% AIXItl
The down-scaled still provably optimal AIXI$tl$ model
\cite[Chp.7]{Hutter:04uaibook} based on universal search
algorithms \cite{Levin:73search,Hutter:02fast,Gaglio:07} was
still computationally intractable.
% OOPS
\indxs{universal}{search}%
The Optimal Ordered Problem Solver \cite{Schmidhuber:04oops}
was the first practical implementation of universal search and
has been able to solve open learning tasks such as Towers-of-Hanoi for
arbitrary number of disks, robotic behavior, and others.

% AIXI Brute Force. 2x2 Matrix Games
For repeated $2\times 2$ matrix games such as the Prisoner's
dilemma, a direct brute-force approximation of AIXI is
computationally tractable. Despite these domains being tiny,
they raise notoriously difficult questions \cite{Shoham:09}. The
experimental results confirmed the theoretical optimality
claims of AIXI \cite{Hutter:06aixifoe}, as far as limited
experiments are able to do so.

% AIXI Monte Carlo
\indxs{Monte-Carlo}{AIXI}%
A Monte-Carlo approximation of AIXI has been proposed in \cite{Pankov:08}
that samples programs according to their algorithmic % makes no sense
probability as a way of approximating Solomonoff's universal
a-priori probability, similar to sampling from
the speed prior \cite{Schmidhuber:02speed}.

% MC-AIXI-CTW
\begin{wrapfigure}{r}{0.62\textwidth} \vspace{-3ex}
\hfill\includegraphics[width=0.6\textwidth]{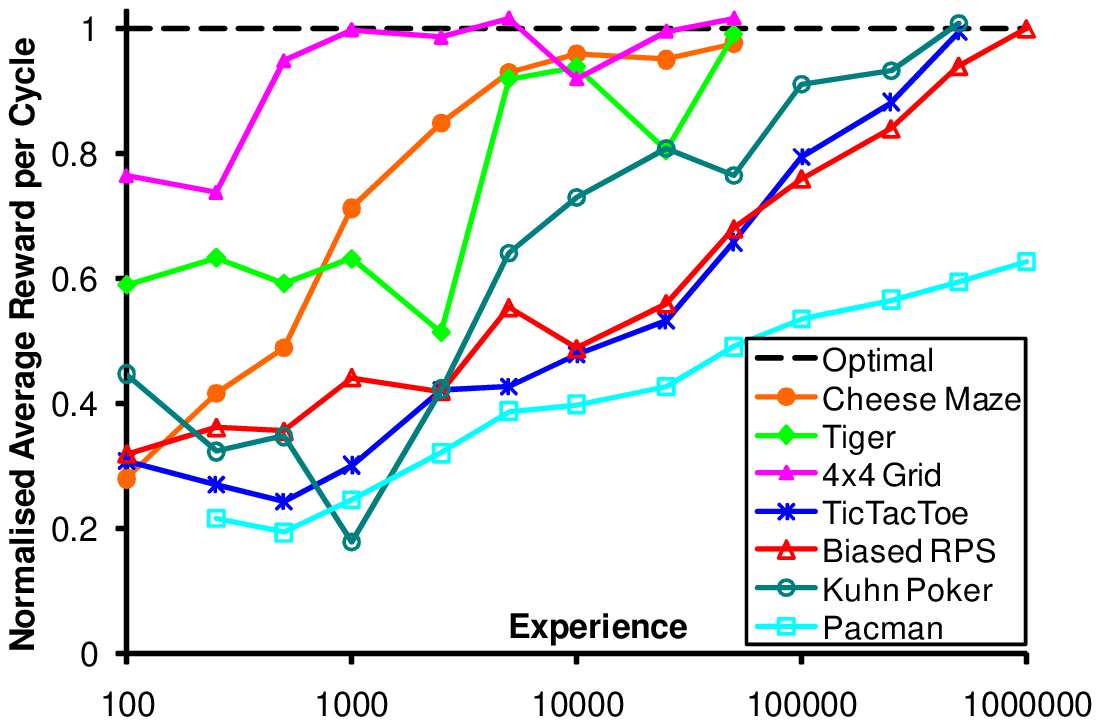}\vspace{-2ex}
\end{wrapfigure}
\indxs{AIXI}{CTW} The most powerful systematic approximation,
implementation, and application of AIXI so far is the
MC-AIXI-CTW algorithm
\cite{Hutter:10aixictw}. It combines
award-winning ideas from universal Bayesian data compression
\cite{Willems:95} and the recent highly successful (in computer
Go) upper confidence bound algorithm for expectimax tree search
\cite{Kocsis:06}. For the first time, without any domain
knowledge, the same agent is able to self-adapt to a
diverse range of environments. For instance, AIXI, is able to
{\em learn} from scratch how to play \idx{TicTacToe}, \idx{Pacman},
Kuhn \idx{Poker}, and other games by trial and error
without even providing the rules of the games \cite{Hutter:11aixictwx}.

%------------------------------%
\paradot{Measures/tests/definitions of intelligence}
%------------------------------%
\indxs{intelligence}{definition}%
\indxs{intelligence}{measure}%
\indxs{intelligence}{test}%
\indxs{Turing}{test}%
The history of informal definitions and measures of intelligence
\cite{Hutter:07idefs} and anthropocentric tests
of intelligence \cite{Turing:50} is long and old.
In the last decade various formal definitions,
measures and tests have been suggested:
Solomonoff induction and Kolmogorov complexity inspired the
universal \idx{C-test}\index{intelligence!C-test}
\cite{Hernandez:00cmi,Hernandez:98}, while AIXI inspired an
extremely general, objective, fundamental, and formal
\indxs{intelligence}{order relation}%
\indxs{intelligence}{universal}%
intelligence order relation \cite{Hutter:04uaibook} and a
universal intelligence measure \cite{Hutter:07iorx,Legg:08},
which have already attracted the popular scientific press
\cite{Fievet:05} and received the SIAI award.
Practical instantiations thereof \cite{Hernandez:10,Legg:11} also
received quite some media attention
(http://users.dsic.upv.es/proy/anynt/).

%------------------------------%
\paradot{Less related/general work}
%------------------------------%
There is of course other less related, less general work,
similar in spirit to or with similar aims as %
UAI/AIXI, e.g.\ %
UTree \cite{McCallum:96}, %
URL \cite{Farias:10}, %
PORL \cite{Suematsu:97,Suematsu:99}, %
FOMDP \cite{Sanner:09}, %
FacMDP \cite{Strehl:07}, %
PSR \cite{Singh:03}, %
POMDP \cite{Doshi:09}, %
and others.
\indxs{feature selection}{reinforcement learning}%
The feature reinforcement learning approach
also belongs to this category
\cite{Hutter:09phimdpx,Hutter:09phidbn,Hutter:10phimp,Hutter:11frlexp}.

%------------------------------%
\paradot{Compression contest}
%------------------------------%
\indxs{compression}{contest}%
\indxs{human}{knowledge}%
The ongoing Human Knowledge Compression Contest
\cite{Hutter:06hprize} is another outgrowth of UAI.
%
% Description
The contest is motivated by the fact that being
able to compress well is closely related to being able to
predict well and ultimately to act intelligently, thus reducing
the slippery concept of intelligence to hard file size numbers.
Technically it is a community project to approximate Kolmogorov
complexity on real-world textual data.
\indxs{intelligent}{compressor}%
\indxs{smart}{compressor}%
In order to compress data, one has to find regularities in
them, which is intrinsically difficult (many researchers live
from analyzing data and finding compact models). So compressors
better than the current ``dumb'' compressors need to be smart(er).
Since the prize wants to stimulate the development of ``universally''
smart compressors, a ``universal'' corpus of data has been
chosen. Arguably the online encyclopedia Wikipedia is a good
snapshot of the Human World Knowledge. So the ultimate
compressor of it should ``understand'' all human knowledge,
i.e.\ be really smart.
%
% Motivation
\indxs{compression}{prize}%
The contest is meant to be a cost-effective way of motivating
researchers to spend time towards achieving AGI via the
promising and quantitative path of compression. The competition
raised considerable attention when launched, but to retain attention
the prize money should be increased (sponsors are welcome),
and the setup needs some adaptation.

%%%%%%%%%%%%%%%%%%%%%%%%%%%%%%%%%%%%%%%%%%%%%%%%%%%%%%%%%%%%%%%
\section{Discussion}\label{sec:Disc}
%%%%%%%%%%%%%%%%%%%%%%%%%%%%%%%%%%%%%%%%%%%%%%%%%%%%%%%%%%%%%%%

%-------------------------------%
\paradot{Formalizing and answering deep philosophical questions}
%-------------------------------%
\indxs{formalization}{intelligence}%
% traits of intelligence
UAI deepens our understanding of artificial (and to a limited extent
human) intelligence; in particular which and how facets of
intelligence can be understood as emergent phenomena of goal- or
reward-driven actions in unknown environments.
%
% UAI allows to formalize and solve philosophical questions
\indxs{philosophy}{intelligence}%
UAI allows a more quantitative and rigorous discussion of
various philosophical questions around intelligence, and
ultimately settling these questions. This can and partly has
been done by formalizing the philosophical
concepts related to intelligence under consideration, and by studying
them mathematically. Formal definitions may not perfectly or
not one-to-one or not uniquely correspond to their intuitive
counterparts, but in this case alternative formalizations allow
comparison and selection.
% UAI may even allow to formalize and solve ethical and social questions
In this way it might even be possible to rigorously answer
various social and ethical questions: whether a super
rational intelligence such as AIXI will be benign to humans and/or
its ilk, or behave psychopathically and kill or enslave humans,
or be insane and e.g.\ commit suicide.

%------------------------------%
\paradot{Building more intelligent agents}
%------------------------------%
\index{agent!building one}\index{building!agents}%
From a practical point of building intelligent agents, since
AIXI is incomputable or more precisely only limit-computable,
it has to be approximated in practice. The results achieved
with the MC-AIXI-CTW approximation are only the beginning.
As outlined in \cite{Hutter:11aixictwx}, many variations and
extensions are possible, in particular to incorporate long-term
memory and smarter planning heuristics.
The same single MC-AIXI-CTW agent is already able to learn to
play TicTacToe, Kuhn Poker, and most impressively Pacman
\cite{Hutter:11aixictwx} from scratch. Besides Pacman, there
are hundreds of other arcade games from the 1980s, and it would
be sensational if a single algorithm could learn them all
solely by trial and error, which seems feasible for (a variant
of) MC-AIXI-CTW. While these are ``just'' recreational games,
they {\em do} contain many prototypical elements of the real
world, such as food, enemies, friends, space, obstacles, objects,
and weapons. Next could be a test in modern virtual worlds
(e.g.\ bots for VR/role games or intelligent software agents
for the internet) that require intelligent agents, and finally
some selected real-world problems.

%-------------------------------%
\paradot{Epilogue}
%-------------------------------%
It is virtually impossible to predict the future rate of
progress but past progress on UAI makes me confident that UAI
as a whole will continually progress. By providing rigorous
foundations to AI, I believe that UAI will also speed up
progress in the field of A(G)I in general. In any case,
UAI is a very useful educational tool with AIXI being
a gold standard for intelligent agents which other
practical general purpose AI programs should aim for.

%%%%%%%%%%%%%%%%%%%%%%%%%%%%%%%%%%%%%%%%%%%%%%%%%%%%%%%%%%%%%%%
%%         Bibliography        %%
%%%%%%%%%%%%%%%%%%%%%%%%%%%%%%%%%%%%%%%%%%%%%%%%%%%%%%%%%%%%%%%

\addcontentsline{toc}{section}{\refname}

\begin{footnotesize}
\newcommand{\etalchar}[1]{$^{#1}$}

\end{footnotesize}

%\addcontentsline{toc}{section}{Index}
%\printindex

\end{document}